\documentclass[letterpaper]{article} 
\usepackage[preprint]{aaai2027} 
\usepackage[hyphens]{url} 
\usepackage{graphicx} 
\usepackage{natbib} 
\usepackage{caption} 
\usepackage{booktabs}
\usepackage{afterpage}

\newcommand{\sysname}{PosterMELD}
\newcommand{\projectlink}{%
  \leavevmode\pdfstartlink attr{/Border[0 0 0]} user{%
    /Subtype/Link/A<< /Type/Action/S/URI/URI(https://github.com/Shannon4Science/PosterMELD) >>}%
  \texttt{Shannon4Science/PosterMELD}\pdfendlink%
}

\title{\sysname: Multi-Agent Paper-to-Poster Generation for Controllable Design Diversity with Editable Print-Ready Outputs}
\author{
Haojie Hu\equalcontrib\textsuperscript{\rm 1,\rm 2},
Chenhao Dang\equalcontrib\textsuperscript{\rm 3,\rm 4},
Yaojia Liu\equalcontrib\textsuperscript{\rm 1,\rm 5},\\
Hengrui Kang\textsuperscript{\rm 3,\rm 4},
Conghui He\textsuperscript{\rm 4},
Weijia Li\textsuperscript{\rm 1}\corresponding
}
\affiliations{
\textsuperscript{\rm 1}Tsinghua Shenzhen International Graduate School, Tsinghua University\\
\textsuperscript{\rm 2}Tongji University\quad
\textsuperscript{\rm 3}Shanghai Jiao Tong University\\
\textsuperscript{\rm 4}Shanghai Artificial Intelligence Laboratory\quad
\textsuperscript{\rm 5}Renmin University of China
}

\begin{document}

\maketitle

\begin{abstract}
Scientific poster construction compresses a long multimodal paper into a readable, editable canvas.
Existing systems hide request-level failures by scoring only completed outputs; direct image generation is not element-editable, while coding-agent workflows are costly.
\sysname{} is a template-conditioned multi-agent pipeline: capacity-aware slots guide writing before rendering, and deterministic gates plus vision--language model (VLM) review route failures to bounded repair.
Each accepted request exports editable PowerPoint (PPTX) and Portable Network Graphics (PNG) artifacts; explicit design controls yield same-paper variants.
Across 621 papers, Print-Ready Rate (PRR) counts requests passing geometric, readability, asset-integrity, and obvious-factual-error checks, with native editability reported separately.
A frozen VLM assigns conditional Craftsmanship--Harmony--Expressiveness (CHE) scores to print-ready outputs.
\sysname{} attains 81.3\% PRR, 3.4 times P2P's rate and 5.2 times PosterGen's, and the highest conditional CHE among generated methods with multiple print-ready outputs.
Native editability and explicit design controls are retained at a mean cost of \$0.38 per request---3.5\% of Codex+Skill's.
Code and resources are available at \projectlink.
\end{abstract}

\section{Introduction}

Scientific posters remain a standard but labor-intensive presentation medium.
Their construction requires reorganizing claims, evidence, and visuals for one large canvas under venue, orientation, and density constraints.
Because revisions continue until printing, native editability and explicit design controls are practical requirements.

Dedicated systems have advanced quickly, from early panel-arrangement and content-extraction models~\cite{qiang2019posters,xu2022posterbot,jaisankar2024postdoc} to multimodal agent pipelines with rendered feedback~\cite{pang2025paper2poster,sun2026p2p,zhang2026postergen,choi2026posterforest}, yet two practical obstacles persist.
The first is \textbf{print-readiness and editability}.
When content is written before the final geometry is fixed, downstream modules must shrink, truncate, or reflow material into incompatible regions, which can produce overflow, overlapping elements, out-of-bounds content, unreadably small text, or missing assets.
Neither prior benchmark in Table~\ref{tab:benchmark-comparison} reports request-level defect rates.
Their evaluations condition aesthetic and content scores on generated outputs, which can omit failed requests from quality averages.
Native editability is also inconsistent across output formats: raster generation flattens figures and tables and prevents element-level revision after export.

\begin{figure}[t]
    \centering
    \includegraphics[width=0.92\columnwidth]{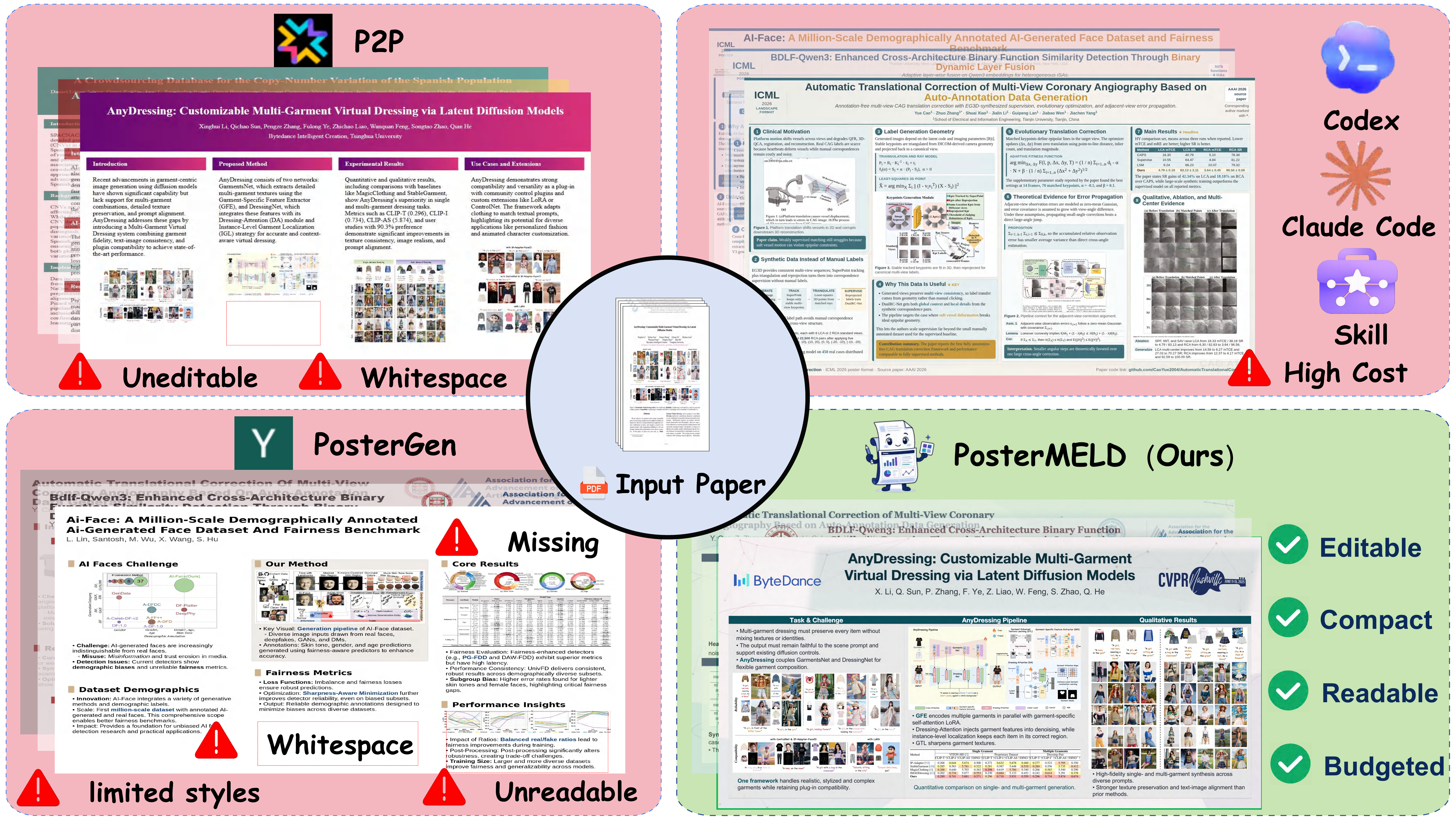}
    \caption{Example outputs for one paper. P2P is non-editable with unfilled regions; PosterGen shows limited layouts and omitted or crowded content; Codex+Skill has the highest request cost (Table~\ref{tab:main-results}); and \sysname{} is editable, compact, and readable. Red markers denote example-specific defects, not aggregate rates.}
    \label{fig:teaser}
\end{figure}

\begin{figure*}[!t]
    \centering
    \includegraphics[width=0.88\textwidth]{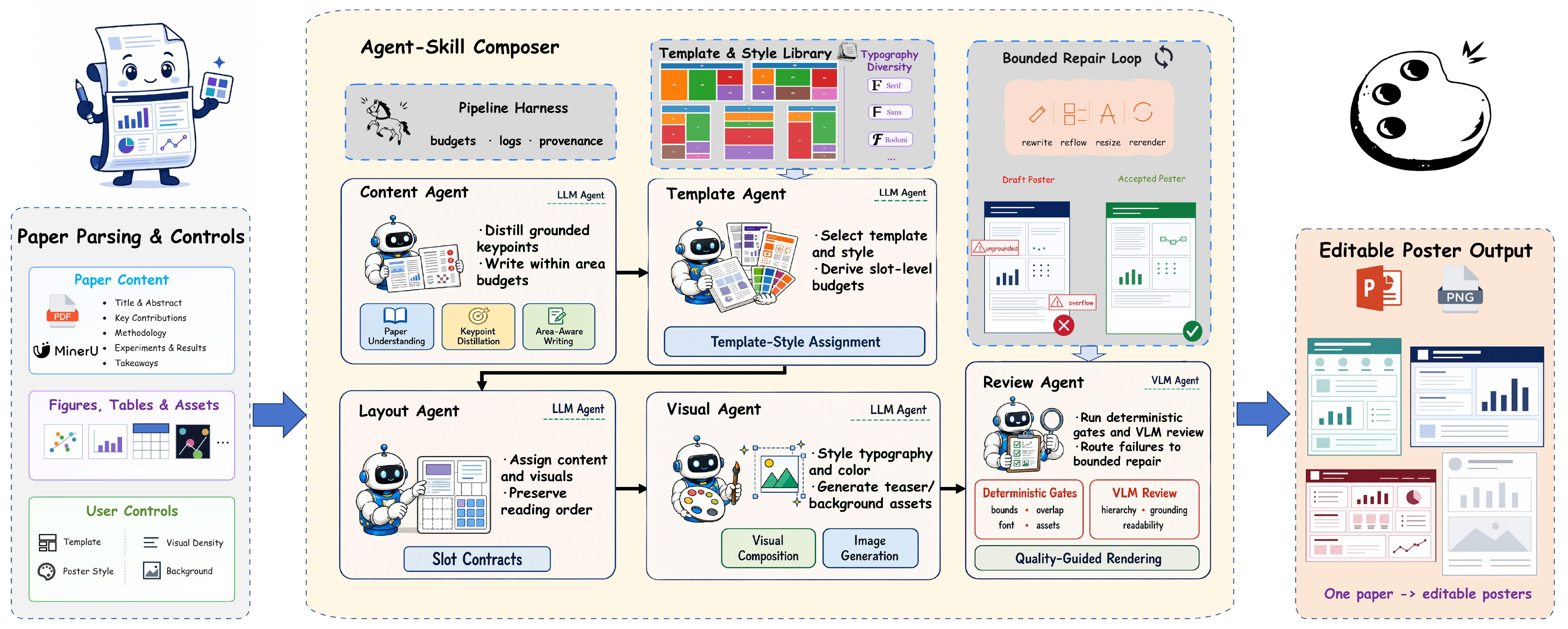}
    \caption{\sysname{}'s Agent-Skill Generation Pipeline. \textbf{Paper Inputs \& Controls} supplies grounded content, source assets, and one explicit control tuple. Inside the \textbf{Agent-Skill Composer}, the \textbf{Content}, \textbf{Template}, \textbf{Layout}, and \textbf{Visual} agents apply embedded skills, while the Template Agent additionally consumes slot contracts from the \textbf{Template \& Style Library}. The \textbf{Review Agent} combines deterministic gates with VLM review; failures enter \textbf{Bounded Repair} and return only to the responsible stage. The \textbf{Pipeline Harness} enforces budgets and records logs and provenance. Each accepted request exports one editable PPTX and its PNG render, while controls are varied across independent requests to obtain controllable design diversity.}
    \label{fig:framework}
\end{figure*}

The second obstacle is \textbf{controllable design diversity}.
Rerunning a stochastic pipeline does not guarantee useful variation: samples may differ only through omitted evidence, tiny text, or broken layouts.
Coding-agent workflows can produce varied posters~\cite{xiao2026researchstudio} but incur high per-artifact cost, whereas direct image generation produces single-shot designs without native editability and with limited text reliability.
Neither paradigm directly provides the combination evaluated here: request-level artifact validity, native editability, explicit control over the design configuration, and measured per-request cost.

\sysname{} integrates multi-agent composition, editable outputs, structural layouts, and controllable design diversity through a template-first principle (Figure~\ref{fig:framework}): structure is fixed before content is written.
A curated library of structural templates exposes each region's capacity through \emph{slot contracts}, so keypoint selection, writing, and visual allocation are \mbox{conditioned} on known geometry rather than squeezed into it afterwards.
Five skill-guided agents compose the poster; each rendered draft then passes deterministic gates and VLM review, and only failed aspects are routed to bounded rewrite, reflow, resize, or rerender actions.
A Pipeline Harness fixes the request configuration and seed, enforces budgets, and records failures, fallbacks, provenance, and cost. Each accepted output is therefore traceable to its control tuple and execution record, and changing those controls across independent requests produces design variants.
To measure validity as well as appearance, a benchmark of 621 papers is assembled---5.1 times the size of the largest released task-specific evaluation set compared here~\cite{sun2026p2p}---and the Print-Ready Rate is evaluated together with aesthetic and content quality, human agreement, and cost under one frozen protocol.

The contributions are threefold:
\begin{itemize}
    \item \textbf{A print-ready multi-agent pipeline for controllable design diversity.} For each request that passes its gates, template-conditioned capacity planning and skill-guided agents produce one editable PPTX file and its PNG render, while a Pipeline Harness couples VLM reflection with deterministic gates and bounded repair and records provenance.
    \item \textbf{A structured template library.} The library contains layout topologies mined from real posters. Their slot contracts---reading order, prominence, character and visual budgets, and asset compatibility---condition writing \emph{before} rendering, while style and density remain orthogonal controls.
    \item \textbf{A large benchmark and a validity-centered protocol.} The largest end-to-end set among the task-specific benchmarks compared here contains 621 papers over 14 \mbox{publication} sources and ten domains. Its request-level \emph{Print-Ready Rate} is reported alongside human and VLM aesthetics, Universal content scores, keypoint fidelity, and cost.
\end{itemize}

\section{Method}

A request supplies \sysname{} with a paper in Portable Document Format (PDF) and one control tuple. If all acceptance gates pass, the request returns one controlled poster as an editable PPTX, its PNG render, and a provenance manifest.
Design variants are requested independently by changing templates, styles, densities, generated-asset settings, or seeds.
The method has two components: a structural template library that makes capacity explicit, and a generation pipeline that plans, writes, reviews, and repairs against that capacity.

\begin{figure}[t]
    \centering
    \includegraphics[width=0.95\columnwidth]{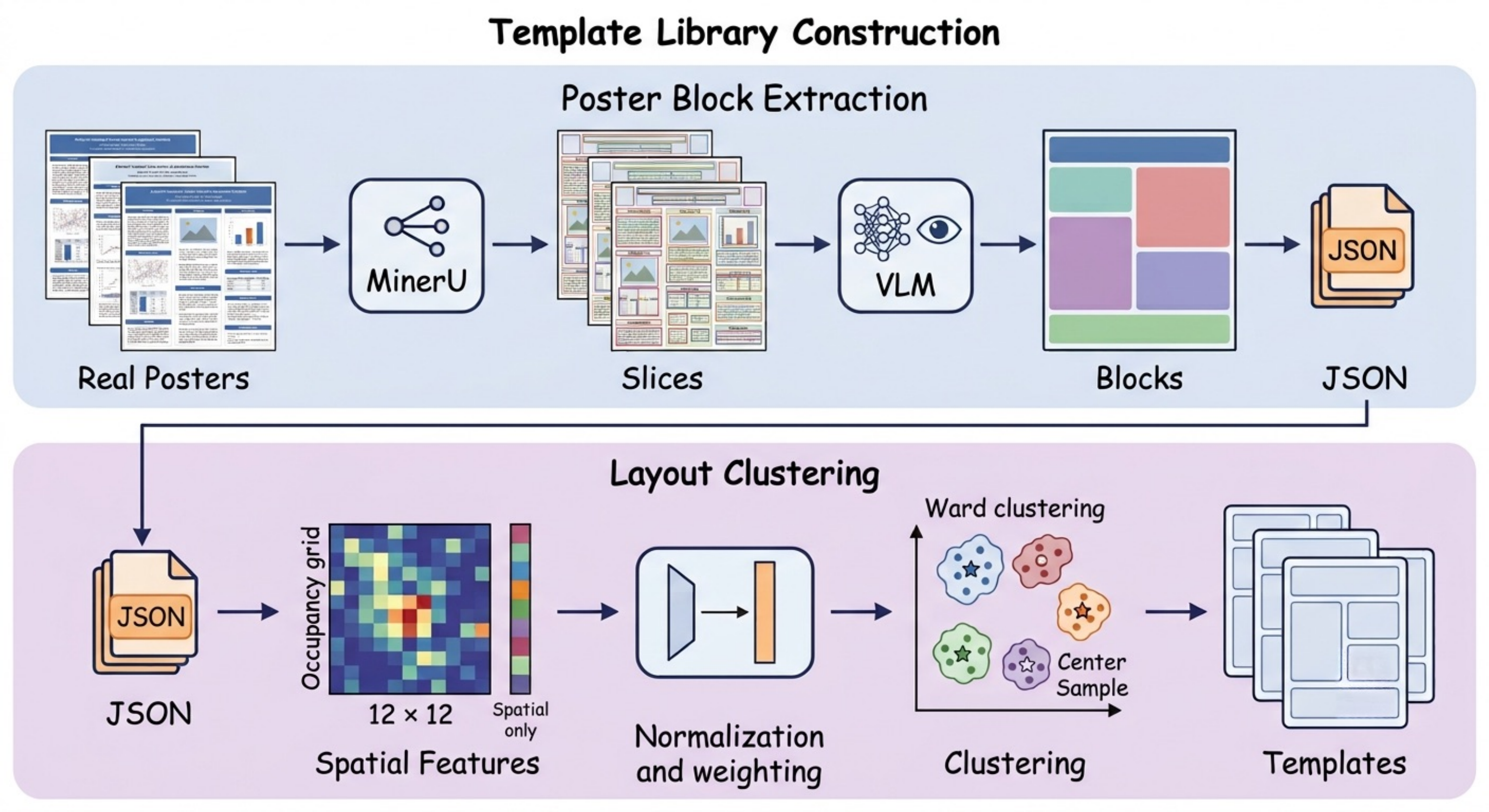}
    \caption{Construction of the template library. Top: author-designed posters are sliced by a MinerU-based extractor and merged by a VLM into semantic blocks under a normalized coordinate system. Bottom: spatial-only layout descriptors built on a $12{\times}12$ occupancy grid are standardized, weighted, and grouped by Ward hierarchical clustering; the sample closest to each cluster center becomes a reusable template.}
    \label{fig:template-library}
\end{figure}

\subsection{Template Library Construction}

The library is mined from author-designed posters in two stages---semantic block extraction and layout clustering (Figure~\ref{fig:template-library})---and enriched into capacity-aware templates.

\paragraph{Poster block extraction.}
A MinerU-based extractor turns each poster into fine-grained slices of text, figures, tables, and equations, each carrying optical character recognition (OCR) content and a bounding box normalized to $[0,1000]$.
Raw slices are too fragmented to serve as layout units, so the poster and all slice coordinates are passed to a vision--language model that merges adjacent, semantically related slices into complete blocks such as Title, Method, Results, and Conclusion under a one-to-one assignment.

\paragraph{Layout clustering.}
The extracted layouts are then clustered using spatial structure only, ignoring noisy semantic labels and text: each poster becomes a fixed-length descriptor covering global geometry, block count, mean block area, a $12{\times}12$ occupancy grid, block shape statistics, estimated row and column structure, and the size ratios of the largest block.
Features are z-score standardized with the occupancy segment up-weighted, Ward hierarchical clustering groups posters into layout families, and the sample closest to each cluster center becomes the representative template.
After validation for legal bounds, non-overlapping regions, minimum region size, coherent reading order, and successful editable rendering, the library contains 24 topologies (16 landscape, 8 portrait).

\paragraph{From templates to slot contracts.}
Each template is augmented with region-level capacity constraints.
The broad top region is identified as the header and the rest ordered top-to-bottom, left-to-right; templates hold 4--10 content slots (mean 6.1).
Every slot stores normalized geometry, area, reading rank, lane membership, and adjacency, augmented by a prominence level, semantic role, target character interval, bullet budget, minimum visual footprint, and text/figure/table compatibility.
Together these fields define a \emph{slot contract} that constrains content selection and writing before rendering.
The library separates structure from presentation: three style profiles govern typography and color, three density profiles govern text and asset budgets, and generated teaser and background assets are independent binary controls, yielding 864 explicit control combinations before seed variation.

\subsection{Agent-Skill Generation Pipeline}


\textbf{Paper Parsing \& Controls} builds a factual representation $F(P)$---title, authors, affiliations, section text, equations, figures, tables, captions, and source locations---with reading order preserved.
The \textbf{Agent-Skill Composer} then advances a \textbf{Shared Typed Poster State} through five agents: four compose a draft, and the fifth wraps rendering, review, and repair in a loop that stops at acceptance or budget exhaustion (Figure~\ref{fig:framework}).
Agents operate on this structured state rather than an unstructured full-paper prompt.
Keypoints retain links to source sections, and every reused asset retains its caption and file provenance for subsequent grounding checks.
Each agent is governed by reusable \emph{skill specifications} that define task instructions, expected state fields, output requirements, validation criteria, and failure handling; the same specifications are reused during repair.

\paragraph{Content Agent.}
The \textbf{Content Agent} turns $F(P)$ into poster-ready material.
The \emph{Paper Understanding} skill interprets the grounded document structure, \emph{Keypoint Distillation} ranks problem, method, evidence, and takeaway units against the aggregate template budget, and \emph{Area-Aware Writing} writes each block within its character and bullet budget, followed by length and grounding checks.

\paragraph{Template Agent.}
The \textbf{Template Agent} uses the \emph{Template-Style Assignment} skill to resolve template and style before any final text is written.
Users may fix the template, style, density, logos, asset generation, and seed; automatic mode selects a compatible template from paper statistics such as section, visual, and table counts, and an incompatible explicit choice is reported rather than silently substituted.
It consumes the \textbf{Template \& Style Library}, exposes each selected template's slot contracts, and combines them with the style and density controls into per-region text, bullet, figure, and table budgets.
This ordering writes content at the target capacity instead of shrinking a fixed summary after layout selection.

\paragraph{Layout Agent.}
The \textbf{Layout Agent} operates on the selected \emph{Slot Contracts}, assigning keypoints and compatible visuals by semantic role and prominence, preserving reading order, and refining geometry so every block respects its declared bounds.
Because assignment follows the slot contract rather than the length of generated text, figures and tables retain their declared minimum footprint.

\afterpage{%
\begin{table}[!t]
    \centering
    {\footnotesize
    \setlength{\tabcolsep}{3.0pt}
    \renewcommand{\arraystretch}{1.12}
    \begin{tabular}{@{}lccccc@{}}
        \toprule
        Benchmark & Papers & Src. & Dom. & Human & Valid. \\
        \midrule
        Paper2Poster & 99 & 3 & 1 & $\times$ & $\times$ \\
        P2PEval & 121 & 9 & 7 & $\times$ & $\times$ \\
        \textbf{\sysname{} (ours)} & \textbf{621} & \textbf{14} & \textbf{10} & $\surd$ & $\surd$ \\
        \bottomrule
    \end{tabular}}
    \caption{Paper-to-poster benchmark comparison. Src.\ and Dom.\ are the numbers of publication-source groups and research domains. Human denotes print-readiness and aesthetic annotation; Valid.\ denotes a defect-free validity metric. Existing paper--poster pairs are reused only as raw examples and are re-annotated.}
    \label{tab:benchmark-comparison}
\end{table}

\afterpage{%
\begin{figure}[!t]
    \centering
    \includegraphics[width=0.98\columnwidth]{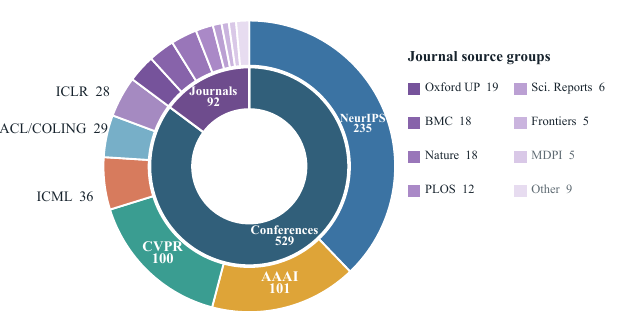}
    \caption{Composition of the 621 paired examples. The inner ring separates 529 conference papers from 92 journal papers; the outer ring resolves six conferences and eight journal source groups.}
    \label{fig:benchmark-composition}
\end{figure}
}
}

\paragraph{Visual Agent.}
The \textbf{Visual Agent} applies the \emph{Visual Composition} skill to typography, color, and logos, emitting text, images, shapes, and tables as native PPTX elements wherever possible and thereby preserving element-level editability.
When enabled, the \emph{Image Generation} tool produces teaser and background assets as presentation aids rather than evidence.
Generation prompts and acceptance gates prohibit measured values and readable pseudo-text in these assets; each generated asset is marked in the manifest, and service failures trigger a recorded deterministic fallback.

\paragraph{Review Agent.}
Structured coordinates alone cannot expose every perceptual failure, so the \textbf{Review Agent} renders each draft before acceptance and examines it with two complementary reviewer classes.
\emph{Deterministic Gates} verify artifact existence and reopenability, PPTX-to-PNG renderability, canvas bounds, section geometry, overlap, lane overflow, minimum text size, asset integrity, and block occupancy.
\emph{VLM Review} inspects the rendered poster and targeted crops for figure legibility, visual hierarchy, grounding, reading order, and conspicuous whitespace or crowding.
The final gate consumes both reports, and a VLM cannot waive a deterministic failure.
The \emph{Quality-Guided Rendering} skill normalizes each issue into a structured record containing its region, severity, category, evidence, and permitted action.
\textbf{Bounded Repair} maps failures to a fixed action set: \emph{rewrite} changes text within the same source facts, \emph{reflow} reassigns or resizes regions, \emph{resize} adjusts typography or visual scale within hard bounds, and \emph{rerender} regenerates a failed visual or the final artifact.
Repair actions are scoped to the failed aspect, limiting changes to previously valid blocks.
Unlike the preceding four agents, the Review Agent runs as a loop and stops when all blocking gates pass or the iteration budget is exhausted; exhausted requests are recorded as failures rather than returned.

\paragraph{Pipeline Harness.}
Around the five agents, the \textbf{Pipeline Harness} freezes the request's configuration and seed, routes structured failures to permitted actions, enforces budgets, validates final artifacts, and records calls, latency, cost, fallbacks, and acceptance state; held-out evaluators play no role during generation.
Requests execute independently, so failures and repairs cannot alter another request's controls.
A passing request yields one \textbf{Editable Poster Output}: a native PPTX, its PNG render, and a provenance manifest.
Independent requests with different controls produce the design variants.

\begin{table*}[!t]
    \centering
    {\footnotesize
    \begin{tabular*}{\textwidth}{@{\extracolsep{\fill}}lccccccccc@{}}
        \toprule
        Method & PRR$\uparrow$ & \multicolumn{4}{c}{Aesthetic Quality (CHE) $\uparrow$} & Universal$\uparrow$ & BERTScore$^{*}$ & Editable & Cost$\downarrow$ \\
        \cmidrule(lr){3-6}
        & & Average & C & H & E & & & & \\
        \midrule
        \multicolumn{10}{@{}l}{\textit{Closed-source systems}} \\[-0.2em]
        GPT-Image-2 & \textbf{85.2} & 2.698 & \textbf{3.028} & \textbf{3.004} & 2.062 & \textbf{4.948} & \textbf{0.804} \tiny \textbf{(+0.010)} & $\times$ & \textbf{0.18} \\
        Codex+Skill & 82.8 & \textbf{2.716} & 3.025 & 3.002 & \textbf{2.121} & 4.876 & 0.814 \tiny (+0.020) & $\surd$ & 10.78 \\
        \midrule
        \multicolumn{10}{@{}l}{\textit{Open-source systems}} \\[-0.2em]
        Paper2Poster & 0.2 & --$^{\dagger}$ & --$^{\dagger}$ & --$^{\dagger}$ & --$^{\dagger}$ & 2.772 & 0.790 \tiny ($-$0.004) & $\surd$ & 0.34 \\
        P2P & 24.2 & 3.071 & 3.280 & 3.027 & 2.907 & 4.033 & 0.829 \tiny (+0.035) & $\times$ & 0.35 \\
        PosterGen & 15.8 & 3.163 & 3.449 & 3.051 & \textbf{2.990} & 3.901 & 0.809 \tiny (+0.015) & $\surd$ & \textbf{0.28} \\
        \textbf{\sysname{}} (ours) & \textbf{81.3} & \textbf{3.247} & \textbf{3.455} & \textbf{3.328} & 2.959 & \textbf{4.456} & \textbf{0.797} \tiny \textbf{(+0.003)} & $\surd$ & 0.38 \\
        \midrule
        Human reference & 98.7 & 3.287 & 3.604 & 3.280 & 2.978 & 4.995 & 0.794 \tiny (--) & $\surd$ & -- \\
        \bottomrule
    \end{tabular*}}
    \caption{Main comparison over 621 papers. PRR (\%) and conditional CHE are measured by the frozen GPT-5.5 judge evaluated in Table~\ref{tab:judge-consistency}. CHE (1--5) averages visual craftsmanship (C), stylistic harmony (H), and expressive distinctiveness (E) over print-ready posters. $^{\dagger}$CHE is not reported for Paper2Poster because only one output is print-ready. Universal assigns zero to missing generations; keypoint-conditioned BERTScore covers successful generations only. Editable requires native figure and text elements, and Cost is the mean U.S. dollars per request including failures. $^{*}$Parentheses report the signed difference between each BERTScore and the human reference. The best result within each source group is \textbf{bold}.}
    \label{tab:main-results}
\end{table*}

\section{Benchmark and Evaluation}

\subsection{Benchmark Composition}

The 621-paper benchmark combines newly curated AAAI 2026 (101), CVPR 2025 (100), and NeurIPS 2025 (200) papers with re-annotated P2PEval (121)~\cite{sun2026p2p} and Paper2Poster (99)~\cite{pang2025paper2poster} pairs.
Among the released task-specific benchmarks compared in Table~\ref{tab:benchmark-comparison}, this is the \textbf{largest end-to-end evaluation set}: 5.1 times the size of P2PEval and 6.3 times the size of the paired Paper2Poster set. Every method is run on all 621 paper requests.

Relative to Paper2Poster and P2PEval, the benchmark expands coverage along three axes.
\emph{Source coverage}: the corpus spans 14 publication-source groups---six conferences plus eight journal groups (Figure~\ref{fig:benchmark-composition})---compared with three and nine for the two existing sets, enabling evaluation across source-specific conventions.
\emph{Domain coverage}: the ten domains extend beyond core artificial intelligence (AI) to biology and medicine, psychology, and the social sciences and include figure-heavy, equation-heavy, and text-heavy papers.
\emph{Supervision structure}: every paper is paired with its author-designed poster; all 1{,}398 posters in the annotation subset carry human print-readiness labels, and the 661 accepted posters carry conditional aesthetic labels---supervision neither existing set provides.
The separate corpus used to mine the template library is disjoint from all benchmark papers and reference posters. A public manifest records source, license, checksum, page count, domain, and pairing, with duplicate and leakage audits.

\subsection{Baselines and Fair Protocol}

Five baselines span three paradigms: \emph{dedicated pipelines}---Paper2Poster~\cite{pang2025paper2poster}, P2P~\cite{sun2026p2p}, and PosterGen~\cite{zhang2026postergen}; \emph{direct image generation}---GPT-Image-2; and \emph{coding-agent stacks}---Codex+Skill, representative of concurrent skill-driven suites~\cite{xiao2026researchstudio}.
GPT-Image-2 and Codex+Skill are closed-source and are reported separately from the open-source pipelines.
Each method receives information derived only from the same source paper; no reference poster is supplied.
Native interfaces are retained: agentic systems ingest the common PDF, whereas GPT-Image-2 receives its MinerU parse and extracted assets.
Artifacts are rendered to a common size, while editability remains separate in Table~\ref{tab:main-results}.
SciPostGen and APEX are excluded as they target layout generation and editing.

A shared configuration is used to reduce implementation variation across systems.
All agentic pipelines---Paper2Poster, P2P, PosterGen, and \sysname{}---run on GPT-4o to reduce backbone variation as a confounder.
Codex+Skill runs on GPT-5.4 with low reasoning effort.
PDF parsing uses the MinerU service throughout template-library construction, benchmark preparation, and the GPT-Image-2 input stage.

\begin{table*}[t]
    \centering
    {\footnotesize
    \begin{tabular*}{\textwidth}{@{\extracolsep{\fill}}lcccccccccc@{}}
        \toprule
        Method & AT & IQ & WS & CR & VT & DA & VC & CF & IF & SE \\
        \midrule
        \multicolumn{11}{@{}l}{\textit{Closed-source systems}} \\[-0.2em]
        GPT-Image-2 & \textbf{4.829} & \textbf{4.936} & \textbf{4.987} & \textbf{4.986} & \textbf{4.911} & \textbf{5.000} & \textbf{4.986} & 4.874 & \textbf{4.997} & \textbf{4.969} \\
        Codex+Skill & 4.792 & 4.763 & 4.850 & 4.982 & 4.736 & 4.987 & 4.897 & \textbf{4.884} & 4.965 & 4.908 \\
        \midrule
        \multicolumn{11}{@{}l}{\textit{Open-source systems}} \\[-0.2em]
        Paper2Poster & 3.401 & 2.585 & 1.773 & 3.150 & 2.140 & 3.448 & 2.335 & 3.543 & 2.483 & 2.866 \\
        P2P & \textbf{4.570} & 3.589 & 3.992 & 4.130 & 3.076 & 4.660 & 3.845 & \textbf{4.147} & 4.151 & 4.164 \\
        PosterGen & 3.396 & 3.713 & 4.116 & 4.279 & 3.477 & 4.528 & 3.729 & 3.626 & 4.166 & 3.976 \\
        \sysname{} (ours) & 3.620 & \textbf{4.240} & \textbf{4.831} & \textbf{4.742} & \textbf{4.327} & \textbf{4.977} & \textbf{4.517} & 4.084 & \textbf{4.789} & \textbf{4.435} \\
        \midrule
        Human reference & 4.992 & 4.998 & 4.984 & 5.000 & 4.981 & 5.000 & 5.000 & 5.000 & 5.000 & 5.000 \\
        \bottomrule
    \end{tabular*}}
    \caption{Universal per-dimension scores over all 621 papers (0--5, missing generations scored zero), following P2P's protocol. AT: authorship/title accuracy; IQ: image uniqueness and quality; WS: balanced white space; CR: contextual relevance; VT: visual-to-text ratio; DA: dimension appropriateness; VC: visual consistency; CF: content fidelity; IF: information flow; SE: self-contained explanation. The best result within each source group is \textbf{bold}. \sysname{} leads the open-source systems on eight of ten dimensions, trailing P2P only on AT and CF, while GPT-Image-2 leads generated methods on nine dimensions. GPT-Image-2 is 0.047 below the human reference on aggregate Universal and 0.589 below it on CHE; the two metrics are therefore interpreted separately.}
    \label{tab:universal}

\end{table*}

\begin{table}[!t]
    \centering
    {\small
    \setlength{\tabcolsep}{4.0pt}
    \begin{tabular}{@{}lcccc@{}}
        \toprule
        Judge & Acc.$\uparrow$ & $\kappa\uparrow$ & $r\uparrow$ & MAE$\downarrow$ \\
        \midrule
        GPT-5.5 & \textbf{0.840} & \textbf{0.683} & \textbf{0.592} & \textbf{0.361} \\
        Gemini-3.1-Pro & 0.810 & 0.622 & 0.485 & 0.420 \\
        Qwen-3.6-27B & 0.810 & 0.623 & 0.291 & 0.449 \\
        Kimi-K2.6 & 0.760 & 0.530 & 0.493 & 0.400 \\
        \midrule
        Human (LOO) & 0.913 & 0.827 & 0.468 & 0.662 \\
        \bottomrule
    \end{tabular}}
    \caption{Agreement of candidate VLM judges with human annotations. PRR accuracy and Cohen's $\kappa$ use 1{,}398 posters; CHE Pearson $r$ and mean absolute error (MAE) use 661 posters accepted by at least three of four annotators. VLM rows are measured against aggregated human targets. Human leave-one-out (LOO) reports each annotator's agreement with the consensus of the other three and is not a performance ceiling. GPT-5.5 is selected before full-benchmark evaluation because it has the strongest agreement among the candidate VLMs.}
    \label{tab:judge-consistency}
\end{table}

\begin{figure*}[t]
    \centering
    \includegraphics[width=\textwidth]{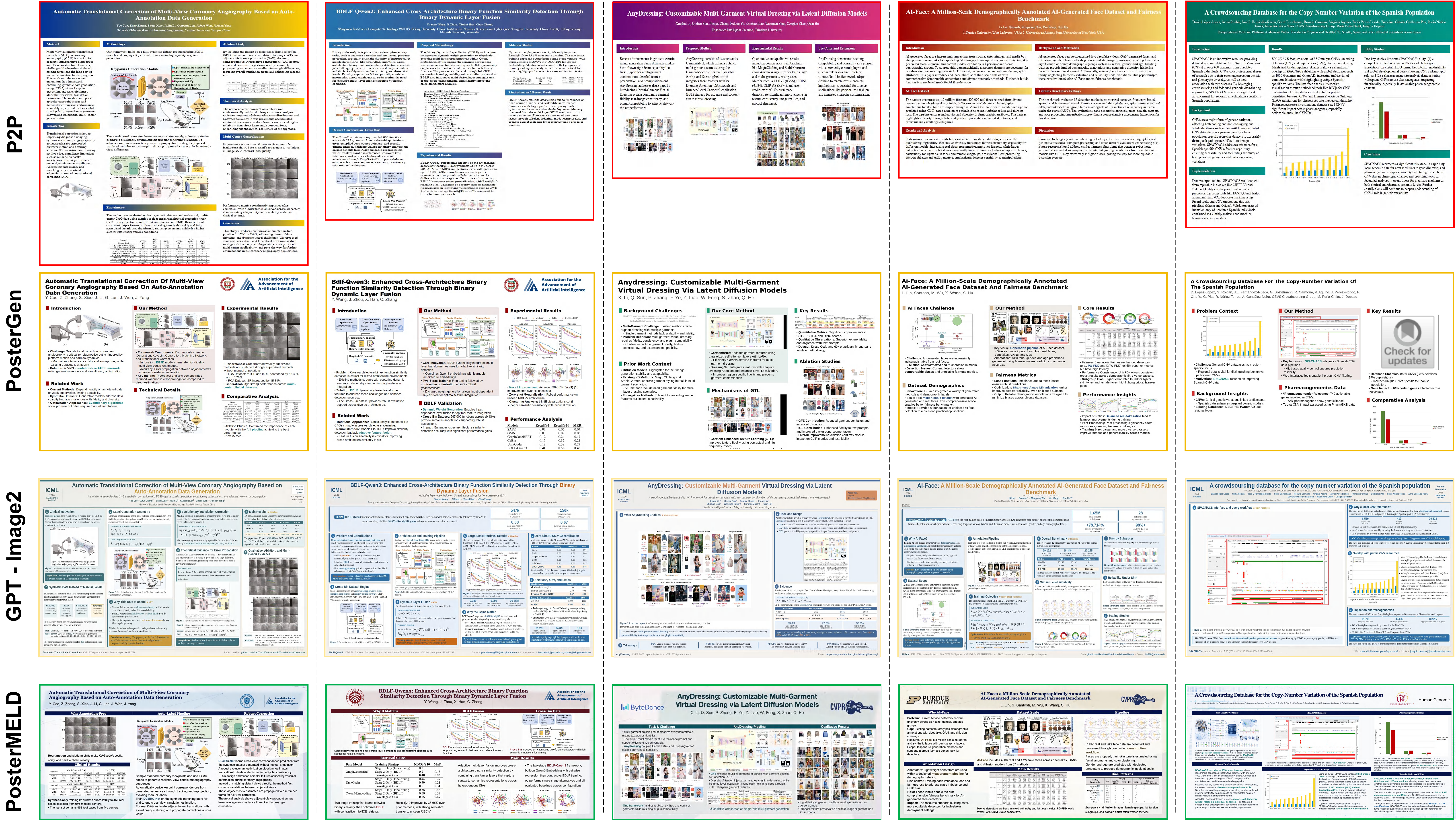}
    \caption{Qualitative comparison on five benchmark papers (columns) across four methods (rows). Methods receive the same paper within each column, and outputs are rendered at a common size. The papers span five research areas. In these examples, P2P and PosterGen use similar layouts across papers, GPT-Image-2 produces non-editable rasters, and \sysname{} varies templates, palettes, and text--figure allocation while preserving editable elements.}
    \label{fig:case-study}
\end{figure*}

\subsection{Print-Ready Rate}

Pipeline completion is separated from output validity.
An output is \emph{print-ready} if an author could print and present it as-is: it must be free of overflow, overlapping elements, out-of-bounds content, unreadably small text, missing or corrupt assets, and confirmed obvious factual errors.
For $N$ requested outputs, let $c_i\in\{0,1\}$ indicate that request $i$ produced the method's declared native artifact and a standardized render, and let $g_i\in\{0,1\}$ indicate that this output is print-ready.
The two rates are
\begin{equation}
\mathrm{Completion}=\frac{1}{N}\sum_i c_i,
\qquad
\mathrm{PRR}=\frac{1}{N}\sum_i c_i g_i,
\label{eq:prr}
\end{equation}
where PRR is the end-to-end fraction of requested posters that both complete and pass every criterion; the shared request-level denominator prevents execution failures from being hidden by a conditional quality score.
Print-readiness is defined by human annotations and measured at scale by a judge calibrated against them.
Four annotators independently label 1{,}398 available renders from a fixed 201-paper subset; print-readiness requires at least three positive labels, while missing renders remain request-level failures.
Human (LOO) compares each annotator with the consensus of the other three and is macro-averaged over four folds; it measures individual-to-consensus agreement rather than a ceiling on VLM-to-consensus agreement.
Table~\ref{tab:judge-consistency} evaluates four candidate VLM judges against these annotations.
GPT-5.5 provides the strongest candidate agreement: 84.0\% accuracy and $\kappa=0.683$ for PRR, together with $r=0.592$ and an MAE of 0.361 for CHE.
It is selected and frozen before evaluation on all 621 papers.

\subsection{Aesthetics, Content, and Cost}

Following the aesthetic evaluation of AI-generated visual content in RealGen~\cite{ye2025realgen}, each rendered poster is scored from 1--5 along three fixed dimensions and summarized by CHE.
\emph{Visual craftsmanship} (C) rates the per-element production quality of figures and text, grounded in facet-based aesthetics and learned image-quality assessment~\cite{moshagen2010facets,talebi2018nima,cao2026artimuse}.
\emph{Stylistic harmony} (H) rates the coherence and attractiveness of the composition, following studies linking objective design factors to subjective perception~\cite{seckler2015linking,an2026aeseval}.
\emph{Expressive distinctiveness} (E) rates design novelty beyond template conventions.
Their mean forms the headline CHE score, with the per-dimension breakdown in Table~\ref{tab:main-results}.
CHE is conditioned on print-readiness; unusable outputs are excluded from CHE and accounted for separately by PRR.
Human ratings are collected under randomized, method-blind presentation on the 661 posters that at least three of four annotators accepted, and the frozen judge applies the identical rubric to the full benchmark.

Content quality follows P2P's published Universal protocol: GPT-4o scores ten dimensions on a 0--5 scale, which are aggregated by a gradient-boosted regressor fitted to human overall judgments (reported $R^2=0.92$)~\cite{sun2026p2p}.
Missing generations are scored zero, so Universal also penalizes completion failures; per-dimension scores appear in Table~\ref{tab:universal}.
Keypoint-conditioned BERTScore is additionally computed over successful generations: poster text is recovered by OCR from the standardized render and compared with ordered paper-derived keypoints using P2P's defaults (RoBERTa-large, layer 17, without inverse-document-frequency weighting or rescaling)~\cite{zhang2020bertscore}.

Cost is averaged in U.S. dollars per requested poster, including all method-internal parsing, generation, image, critique, repair, and failed-attempt spending while excluding held-out evaluation.
This request-level denominator includes failed attempts and remains defined when a method produces no print-ready poster.

\section{Experiments}

\subsection{End-to-End Comparison}

Across 621 requests, \sysname{} attains 81.3\% PRR, compared with 24.2\% for P2P, 15.8\% for PosterGen, and 0.2\% for Paper2Poster (Table~\ref{tab:main-results}).
This result is within 3.9 percentage points of GPT-Image-2 and 1.5 points of Codex+Skill; using unrounded rates, it is 3.4 times the PRR of P2P and 5.2 times that of PosterGen.
\sysname{} has the highest conditional CHE (3.247) among generated methods with at least two print-ready outputs.
Its craftsmanship score (3.455) is closest to the human reference, with a gap of 0.149 points.

Universal and BERTScore produce a different ordering from PRR and CHE.
GPT-Image-2 scores 4.948 on Universal, 0.047 below the human reference, and attains 85.2\% PRR, but its non-editable raster has the lowest conditional CHE, 2.698, among methods with at least two print-ready outputs.
BERTScore ranks P2P at 0.829 above the human reference at 0.794 because it measures text-to-keypoint similarity rather than artifact validity or aesthetics.
\sysname{} leads the open-source systems on Universal and on eight of its ten dimensions.
Universal and BERTScore are therefore reported as content diagnostics; PRR measures artifact validity, and CHE measures aesthetics conditional on validity.

Artifact validity, editability, aesthetic quality, and cost remain separate evaluation axes.
Among generated methods, GPT-Image-2 has the lowest cost at \$0.18 and the highest PRR at 85.2\%, but does not provide editable output.
Codex+Skill is editable and reaches 82.8\% PRR, with a conditional CHE of 2.716 at \$10.78 per request.
\sysname{} is editable and reaches 81.3\% PRR and 3.247 conditional CHE at \$0.38 per request, compared with \$0.35 for P2P; the mean request cost of Codex+Skill is approximately 28 times that of \sysname{}.
Figure~\ref{fig:case-study} provides a qualitative comparison of the displayed artifacts: P2P and PosterGen use similar layouts across the five examples, GPT-Image-2 produces non-editable rasters, and \sysname{} uses different templates, palettes, and text--figure allocations.

\subsection{Qualitative Analysis of Design Controls}

Controllable design diversity is examined qualitatively by generating multiple poster variants from the same scientific paper.
Although all variants are grounded in a shared source, they can differ in overall visual presentation, organization, emphasis, and stylistic expression.
This one-paper-to-many-posters setting illustrates the range of design outcomes supported by the pipeline beyond a single fixed layout, while keeping the underlying research topic and evidence consistent.
Additional examples and complete qualitative results are provided in the Supplementary Document.

\subsection{Limitations}

The pipeline depends on external foundation models and parsers, so parsing errors on unusual layouts propagate into the poster, and automatic gates certify geometry and legibility rather than scientific correctness, leaving final verification to the authors.
The frozen judge may share biases with the generators; calibration against human consensus and comparison across four judge families quantify but do not eliminate this risk.
The mined library covers common academic structures rather than every design culture, portrait topologies remain less mature than landscape ones, and generated assets carry hallucination risk despite provenance marking.
Finally, CHE is conditioned on print-readiness and must be read jointly with PRR, and the diversity analysis is qualitative, so its claims cover observed control-conditioned changes rather than a scalar diversity gain.

\section{Related Work}

\paragraph{Automated scientific poster generation.}
Scientific-poster systems have progressed from panel layout and content selection~\cite{qiang2019posters,xu2022posterbot,jaisankar2024postdoc} to editable planning, agent--checker pipelines, aesthetic optimization, and post-hoc editing~\cite{pang2025paper2poster,sun2026p2p,zhang2026postergen,choi2026posterforest,shi2026apex}.
Poster-specific models~\cite{tanaka2024scipostlayout,zhong2025scientific,inadumi2026scipostgen,hsu2023posterlayout} coexist with diverse document-layout generation~\cite{kang2025omnidoclayout} and general constrained-layout synthesis~\cite{li2019layoutgan,lee2020ndn,kong2022blt,inoue2023layoutdm,zhang2023layoutdiffusion,lin2023layoutprompter,feng2023layoutgpt}.
Evaluation nevertheless remains output-conditional; request-level validity, faithful content, native editability, and design control are not assessed jointly.

\paragraph{Agentic artifacts and quality control.}
Document layout analyzers recover structural regions from heterogeneous pages~\cite{zhao2024doclayout}, while slide-generation systems summarize sources, plan multimodal slides, reconstruct editable designs, and manipulate presentation objects~\cite{sun2021d2s,fu2022doc2ppt,bandyopadhyay2024slides,tang2025slidecoder,zheng2025pptagent,xu2025pregenie,jung2026talk}.
Synthetic-image supervision improves open-source image generation~\cite{ye2025echo4o}; agentic visual generation adds cognitive search and reasoning or code-mediated canvases~\cite{he2026mindbrush,ye2026genclaw}, while verbal reflection provides a general mechanism for iterative refinement~\cite{shinn2023reflexion}.
Concurrent poster systems add VLM repair, deterministic figure insertion, and cross-artifact coding skills~\cite{vinaykumar2026any2poster,yang2026posterharness,xiao2026researchstudio}.
The contribution lies in their integration with template-first capacity planning, bounded repair, explicit design controls, and request-level print-readiness evaluation.

\section{Conclusion}

\sysname{} is a template-first multi-agent pipeline for generating editable, print-ready scientific posters under explicit design controls.
Capacity-aware slot contracts guide five skill-specialized agents within fixed geometry; deterministic gates, VLM review, and bounded repair enforce artifact validity while preserving editability.
Each accepted request produces a native PPTX, matched PNG render, and provenance record, while independent control settings provide design variation.
On 621 papers, \sysname{} achieves 81.3\% PRR---3.4 times P2P and 5.2 times PosterGen---and the highest conditional CHE among generated methods (3.247), at \$0.38 per request.
These results show that print readiness, native editability, and controllable diversity can be achieved jointly in practical paper-to-poster generation.

\bibliography{poster_references}

\end{document}